\documentclass[sigconf]{acmart}

\usepackage{graphicx}
\usepackage{subcaption}
\usepackage{tabularx}
\usepackage{multirow}
\usepackage{float}

\AtBeginDocument{%
  }

\copyrightyear{2025}
\acmYear{2025}
\setcopyright{acmlicensed}\acmConference[MMAsia '25]{ACM Multimedia Asia}{December 9--12, 2025}{Kuala Lumpur, Malaysia}
\acmBooktitle{ACM Multimedia Asia (MMAsia '25), December 9--12, 2025, Kuala Lumpur, Malaysia}
\acmDOI{10.1145/3743093.3770984}
\acmISBN{979-8-4007-2005-5/2025/12}

\begin{document}

\title{APGNet: Adaptive Prior-Guided for Underwater Camouflaged Object Detection}


\author{Xinxin Huang}
\affiliation{%
  \institution{Nanjing University of Aeronautics and Astronautics}
  \city{Nanjing}
  \country{China}
}
\email{xxhnuaa@163.com}

\author{Han Sun}
\authornote{Corresponding author.}
\affiliation{%
  \institution{Nanjing University of Aeronautics and Astronautics}
  \city{Nanjing}
  \country{China}
}
\email{sunhan@nuaa.edu.cn}

\author{Junmin Cai}
\affiliation{%
  \institution{Nanjing University of Aeronautics and Astronautics}
  \city{Nanjing}
  \country{China}
}
\email{cjm\_nuaa@nuaa.edu.cn}

\author{Ningzhong Liu}
\affiliation{%
  \institution{Nanjing University of Aeronautics and Astronautics}
  \city{Nanjing}
  \country{China}
}
\email{lnz\_nuaa@163.com}

\author{Huiyu Zhou}
\affiliation{%
  \institution{University of Leicester}
  \city{Leicester}
  \country{UK}
}
\email{hz143@leicester.ac.uk}



\renewcommand{\shortauthors}{Huang et al.}

\begin{abstract}
 Detecting camouflaged objects in underwater environments is crucial for marine ecological research and resource exploration. However, existing methods face two key challenges: underwater image degradation, including low contrast and color distortion, and the natural camouflage of marine organisms. Traditional image enhancement techniques struggle to restore critical features in degraded images, while camouflaged object detection (COD) methods developed for terrestrial scenes often fail to adapt to underwater environments due to the lack of consideration for underwater optical characteristics.
 To address these issues, we propose APGNet, an Adaptive Prior-Guided Network, which integrates a Siamese architecture with a novel prior-guided mechanism to enhance robustness and detection accuracy. First, we employ the Multi-Scale Retinex with Color Restoration (MSRCR) algorithm for data augmentation, generating illumination-invariant images to mitigate degradation effects. Second, we design an Extended Receptive Field (ERF) module combined with a Multi-Scale Progressive Decoder (MPD) to capture multi-scale contextual information and refine feature representations. Furthermore, we propose an adaptive prior-guided mechanism that hierarchically fuses position and boundary priors by embedding spatial attention in high-level features for coarse localization and using deformable convolution to refine contours in low-level features.
 Extensive experimental results on two public MAS datasets demonstrate that our proposed method APGNet outperforms 15 state-of-art methods under widely used evaluation metrics. 
\end{abstract}




\maketitle

\begin{figure}[h]
  \centering
  \includegraphics[width=\linewidth]{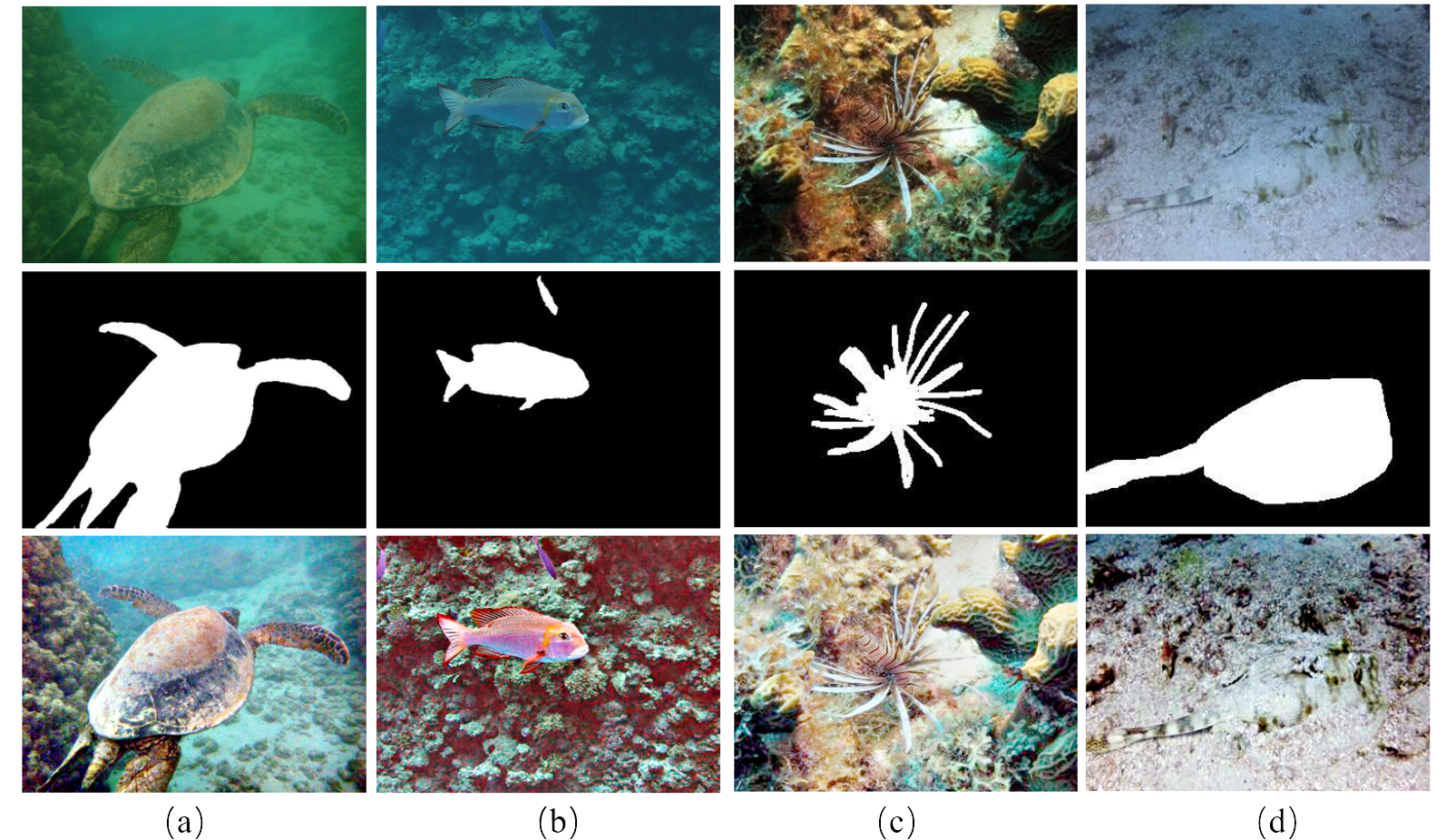}
  \caption{Examples of underwater images.The first row displays the original images, the second row shows the ground truth images, and the third row presents the images processed by the MSRCR algorithm.}
  \label{fig1}
\end{figure}
\section{Introduction}
The exploration of underwater environments is pivotal for advancing research in marine ecology, geology, and resource exploration. Recent advancements in underwater robotics and intelligent monitoring systems have enabled significant progress, driven by sophisticated image processing algorithms for tasks such as scene parsing and object recognition. However, segmenting marine animals in underwater environments remains a formidable challenge due to the inherent complexity of underwater imagery. Specifically, light absorption and scattering effects degrade image quality, leading to poor visibility, low contrast, and color distortion. These challenges are further exacerbated when marine animals exhibit natural camouflage, blending seamlessly with their surroundings—a survival trait that complicates accurate segmentation. For instance, as illustrated in Fig. \ref{fig1}(a–b), degraded underwater images obscure critical details, while Fig. \ref{fig1}(c–d) highlights camouflaged marine organisms that are visually indistinguishable from their backgrounds.

Existing approaches for underwater image enhancement, such as restoration techniques \cite{ancuti2012enhancing}, often fail to address the unique demands of marine animal segmentation (MAS). While camouflaged object detection (COD) methods focus on identifying objects integrated with backgrounds, they are primarily designed for terrestrial scenarios and neglect underwater-specific degradation factors like light scattering and wavelength-dependent absorption. Consequently, these methods struggle to generalize to underwater environments, where both image quality degradation and object camouflage coexist.

To overcome these limitations, we propose APGNet, an Adaptive Prior-Guided Underwater Camouflaged Object Detection algorithm. Our framework integrates a Siamese network architecture with data augmentation techniques to mitigate the effects of image degradation. By employing the Multi-Scale Retinex with Color Restoration (MSRCR) \cite{MSRCR} for image enhancement, we generate robust inputs that preserve critical features under diverse underwater conditions. To address the challenge of capturing multi-scale contextual information, we introduce an Extended Receptive Field (ERF) module, which combines dilated and asymmetric convolutions to enhance the network’s ability to discern camouflaged objects in cluttered backgrounds. Furthermore, we design an adaptive prior-guided strategy that dynamically fuses position and boundary priors across network stages. Low-level features are refined using boundary priors to sharpen edges, while high-level features leverage position priors for coarse localization, enabling precise segmentation through progressive refinement. The network is optimized via a hybrid loss function combining task-specific segmentation loss and feature alignment loss, ensuring both accuracy and generalization. Experimental results demonstrate that APGNet significantly outperforms existing COD methods in underwater scenarios, achieving state-of-the-art performance in marine animal segmentation.

In summary, our contributions are as follows:  
\begin{itemize}
    \item A Siamese-based framework enhanced by MSRCR preprocessing, tailored to address underwater degradation and camouflage.
    \item An adaptive prior fusion mechanism that hierarchically integrates position and boundary cues, balancing local details and global context.
    \item Comprehensive validation of benchmark datasets, demonstrating superior robustness and segmentation accuracy under challenging underwater conditions.
\end{itemize}

\section{Related work}
\subsection{Marine Animal Segmentation}
Marine Animal Segmentation (MAS) is a challenging task in image segmentation, requiring solutions for underwater-specific issues like light absorption, scattering, low contrast, and camouflage, which obscure object boundaries. Early CNN-based methods, such as ECDNet \cite{ECDNet}, enhance feature extraction via cascade decoders but struggle with diverse water conditions. To address this, Chen et al. \cite{Siamese} propose a Siamese network to mitigate degradation effects, while Cheng et al. \cite{cheng2023bidirectional} leverage structural texture and contextual clues through bidirectional mentoring. However, these methods lack global perception due to CNNs’ local receptive fields. Recent Vision Transformers (ViTs) \cite{vit} improve long-range dependency modeling but require extensive labeled data. Hybrid approaches (e.g., Hong et al. \cite{hong2023usod10k}) combine CNNs and ViTs but remain limited in handling complex camouflage.
Different with privious works, our method diverges by integrating a vision foundation model with adaptive prior guidance, achieving robust segmentation in challenging underwater scenes with limited annotations.

\subsection{Camouflaged Object Detection}
Camouflaged Object Detection (COD) focuses on identifying objects that blend seamlessly into their surroundings, presenting significant challenges due to the intrinsic similarity between objects and backgrounds. 
The rise of deep learning has driven substantial progress in COD. Fan et al. \cite{fan2021concealed} pioneered a large-scale dataset (COD10K) with 10,000 images across 78 categories and introduced a baseline network using receptive field expansion and partial decoding. Le et al. \cite{le2019anabranch} further enriched the field with the CAMO dataset and a multi-task framework combining classification and segmentation.
Recent innovations focus on advanced feature learning, where Li et al. \cite{li2021uncertainty} exploited contradictory cues between salient and camouflaged regions, Yang et al. \cite{yang2021uncertainty} fused Bayesian uncertainty with Transformer architectures for boundary reasoning, and Pang et al. \cite{pang2022zoom} mimicked human visual search through mixed-scale triplet learning.
Despite advancements, COD struggles in extreme environments (e.g., underwater scenes) where degradation and homogeneous backgrounds exacerbate camouflage effects. This motivates our work to integrate degradation-adaptive mechanisms and hierarchical prior guidance for robust detection.

\section{Our Method}
\subsection{Overall Network Framework}
\begin{figure}[h]
  \centering
  \includegraphics[width=\linewidth]{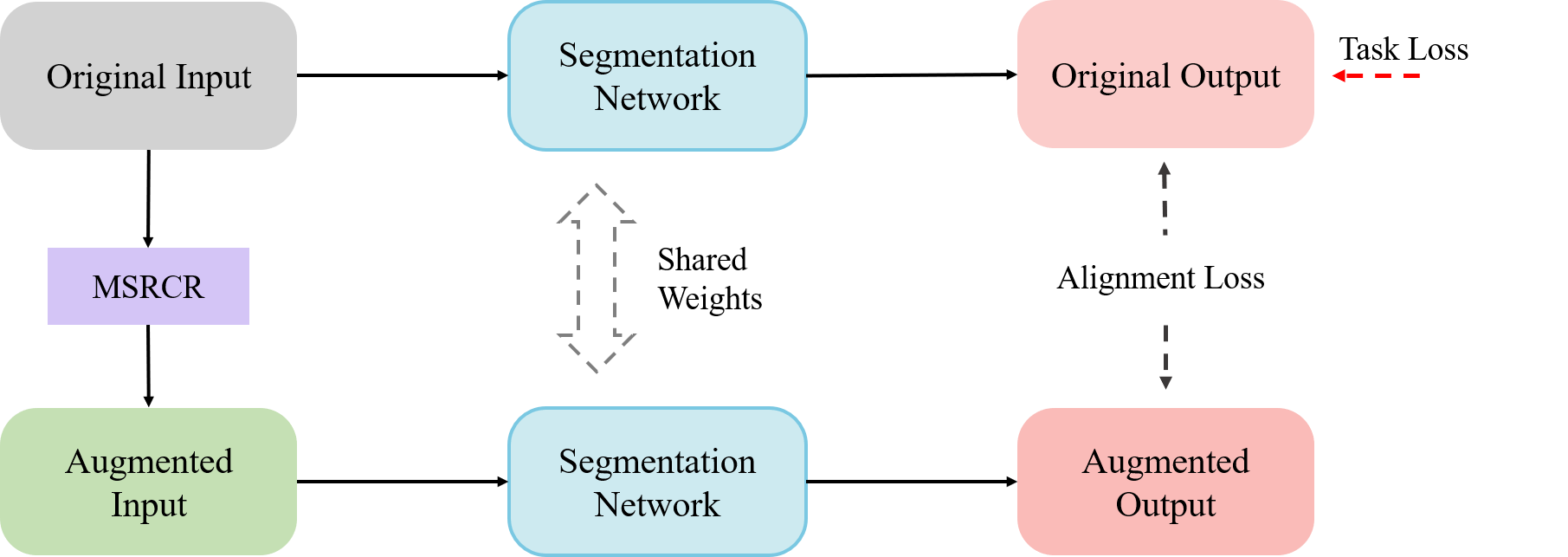}
  \caption{Siamese network architecture.} \label{fig2}
\end{figure}

Fig.\ref{fig2}  illustrates the overall framework, comprising a data augmentation module (MSRCR), a Siamese segmentation network, and a hybrid loss function. During training, the augmentation module enhances data diversity and model robustness by generating transformed instances. The Siamese segmentation network processes both original and augmented images, producing two sets of predictions. Task loss optimizes segmentation accuracy, while alignment loss enforces consistency between predictions, enabling the network to learn shared semantic features and improving adaptability to underwater environments.

\subsection{Data Augmentation Strategy for Camouflage and Degradation}
Underwater environments pose two major challenges for marine animal segmentation, which include image degradation due to suspended particles, reduced brightness in deep-sea conditions, and color distortion in turbid waters, along with the natural camouflage of marine organisms. To tackle these problems, we adopt the Multi-Scale Retinex with Color Restoration algorithm \cite{MSRCR}, which improves image quality by decomposing illumination and reflection components through multi-scale filtering. The core operation is expressed as follows.
\begin{equation}
        R_{MSRCR} = \sum_{n=1}^{n_{s}}\lambda_{n}\left\{\lg I-\lg[I\cdot G_{n}(x,y)]\right\}\cdot C_i
\end{equation}
where $G_{n}$ is a Gaussian filter at scale $n$, and $C_i$ is a color balance adjustment factor. As shown in Fig.\ref{fig1}, MSRCR significantly improves contrast while preserving details in both dark and bright regions, providing high-quality inputs for downstream segmentation tasks.

\begin{figure*}[!htbp]
  \centering
  \includegraphics[width=\textwidth]{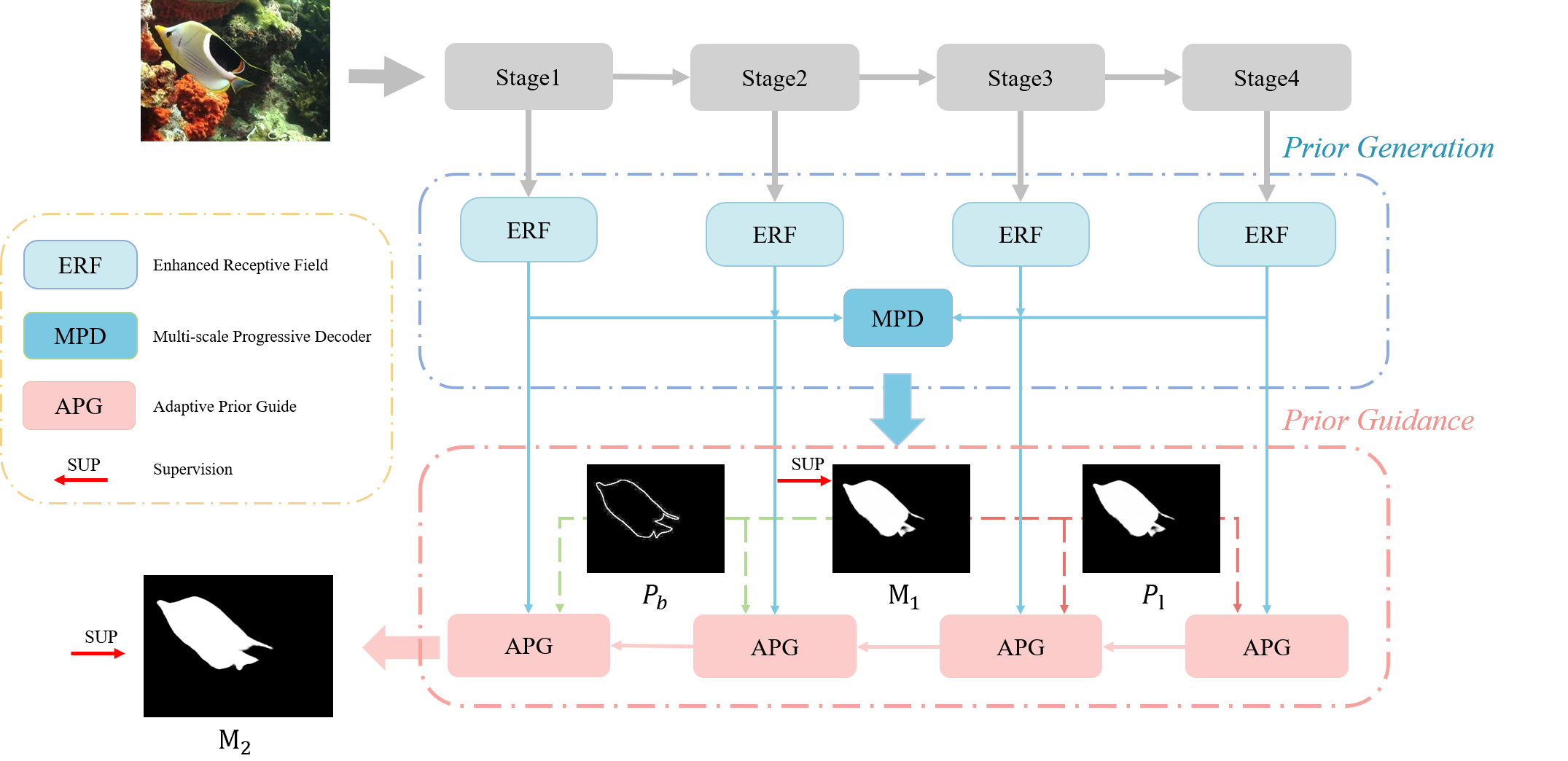}
  \caption{The overall architecture of the proposed APGNet.}
  \label{fig3}
\end{figure*}
\subsection{Segmentation Network Framework}
As shown in Fig.\ref{fig3}, the proposed segmentation network, APGNet, consists of two main stages, namely the prior generation stage and the prior-guided stage. In the prior generation stage, a rough prediction of camouflaged objects is obtained using the extended receptive field (ERF) module and a multi-scale progressive decoder(MPD), which extract position and boundary priors for subsequent fine segmentation. The prior-guided stage then incorporates these priors into object features via the adaptive prior-guided module, progressively refining the segmentation and producing a detailed prediction of the camouflaged object. The following sections will elaborate on the design and implementation details of each key module.

\subsection{Prior Generation Stage}
\subsubsection{Extended Receptive Field Module (ERF)}
To enhance the multi-scale contextual information extracted by PVTv2, we introduce the extended receptive field module (ERF)
, inspired by the characteristics of the human visual system. Research indicates that receptive fields of different sizes play a crucial role in object detection. To leverage this, ERF combines dilated convolution and asymmetric convolution techniques to effectively capture multi-scale contextual information while expanding the receptive field. Dilated convolution enlarges the receptive field without increasing computational cost, enabling the network to aggregate information from a broader area. Asymmetric convolution further enhances adaptability to complex structures by applying convolutions in multiple directions. This combined strategy significantly improves the detection accuracy of camouflaged objects, particularly in challenging environments. The process is as follows:
\begin{equation}
    {f_i}' = Conv(Concat\{ {C_{dil}}({f_i}),{C_{asy}}({f_i})\} )
\end{equation}

\begin{figure}[H]
  \centering
  \includegraphics[width=\linewidth]{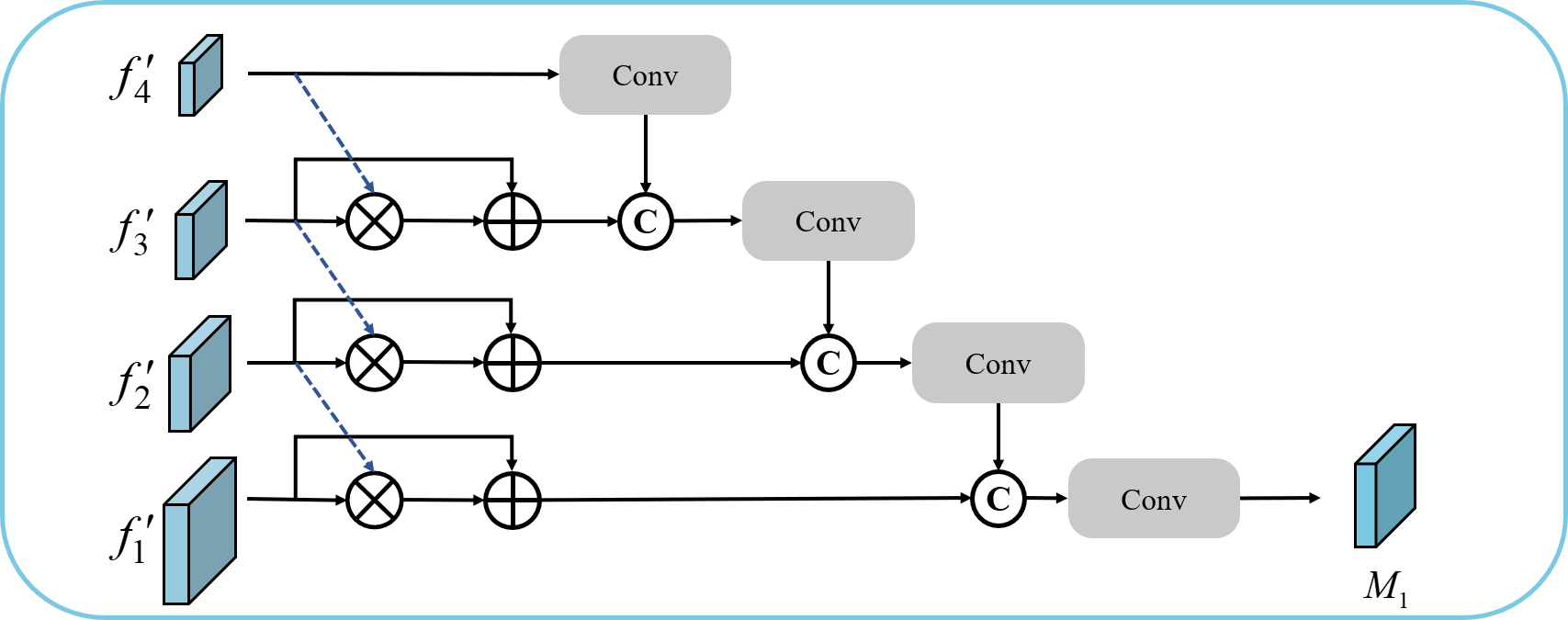}
  \caption{The detailed architecture of the Multi-Scale Progressive Decoder.} \label{fig4}
\end{figure}
\subsubsection{Multi-Scale Progressive Decoder (MPD)}
Scale variation is a major challenge in underwater camouflaged object detection, as small objects demand fine details while large ones require global contours. Traditional feature concatenation often fails to fully exploit deep semantic information, limiting accuracy in complex environments. To address this, we propose a Multi-Scale Progressive Decoder (MPD) that dynamically integrates deep features as semantic filters to suppress background noise and emphasize salient regions, while residual connections preserve original details, as shown in Fig.\ref{fig4}. In our framework, multi-level feature maps \(F' = \{ f'_i \}_{i=1}^{4}\) are progressively refined. the deepest feature \(f'_4\) is enhanced via a 3\(\times\)3 convolution, and each shallower feature \(f'_i\) is filtered by \(f'_{i+1}\) to extract high-level cues. The final refined feature \(\hat{f}'_1\) yields an initial prediction map \(M_1\) as follows:
\begin{equation}
\hat{f}'_{i} = \text{Conv}\Big( \text{Concat}\big(\hat{f}'_{i+1},\, f'_{i} \otimes f'_{i+1} \oplus f'_{i}\big)\Big)
\end{equation}
where \(\text{Conv}()\) denotes a 3\(\times\)3 convolution, \(\text{Concat}()\) represents channel concatenation, and \(\oplus\) is element-wise summation.

The rough prediction \(M_1\) is passed through a Sigmoid function to generate a positional prior \(P_l\) for early detection, while a fixed convolution extracts a boundary prior \(P_b\) to refine object contours. This integrated approach effectively combines spatial and structural cues to enhance segmentation accuracy.

\begin{figure}[H]
  \centering
  \includegraphics[width=\linewidth]{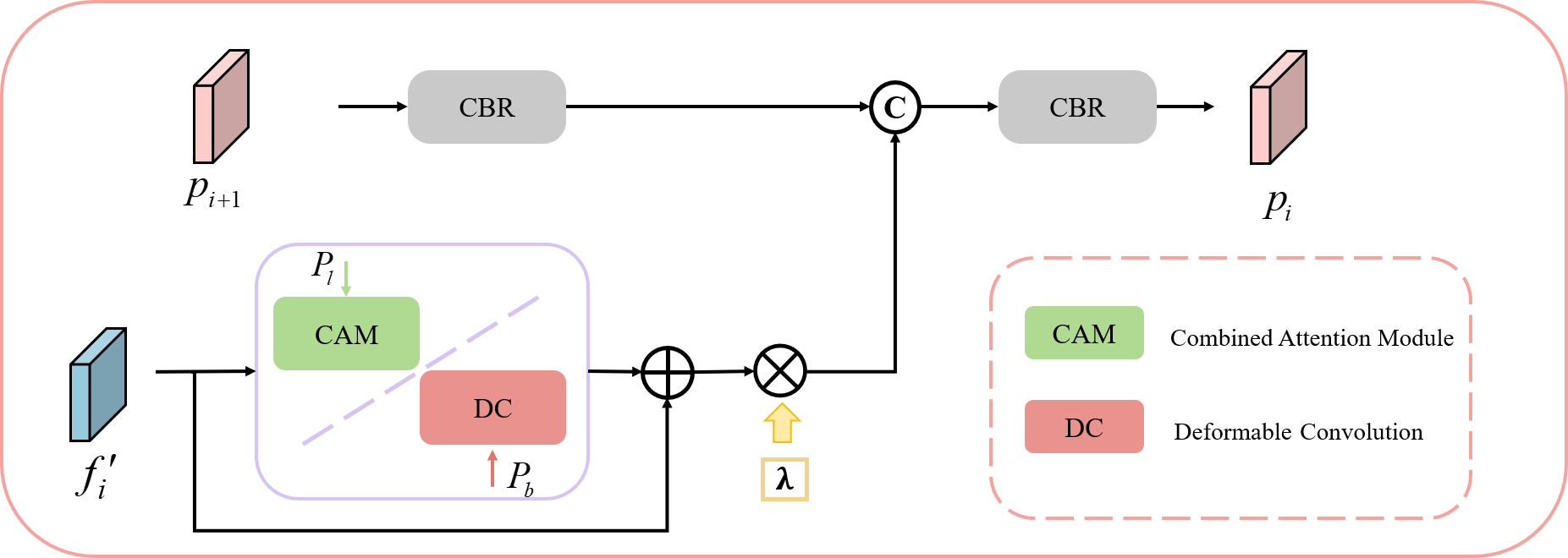}
  \caption{The detailed architecture of the Adaptive Prior-Guided Module.} \label{fig5}
\end{figure}
\subsection{Prior-Guided Stage}
Prior information plays a crucial role in refining feature representations at different network stages in camouflaged object segmentation. High-level features primarily capture semantic context and benefit from position priors for coarse localization, while low-level features focus on fine details and require boundary priors for precise edge refinement. To address this, we propose an Adaptive Prior-Guided (APG) Module that dynamically integrates prior information into feature learning. The detailed architecture is shown in Fig.\ref{fig5}.

Specifically, the APG module processes multi-scale features \(F' = \{ f'_i \}_{i=1}^{4}\) and prior information \(P\) using a learnable fusion weight \(\lambda\):
\begin{equation}
    f_i^{guided} = \lambda  \cdot ({\rm{f}}({f'_i},P) \oplus {f'_i})
\end{equation}
where 
$\rm{f}$ is a function that varies depending on  the type of prior. Position priors are incorporated via a Combined Attention Module (CAM) that applies spatial and channel-wise attention to enhance feature selectivity, while boundary priors are introduced through Deformable Convolution (DC) to refine edge structures. The guided features are then progressively fused with deeper layers, yielding a final segmentation output $M_2$ with enhanced structural accuracy and fine detail preservation.

\subsection{Loss Function}
The total loss consists of a segmentation loss and an alignment loss. The segmentation loss combines weighted binary cross-entropy and IoU loss to address class imbalance and enhance spatial overlap for both rough and final predictions. The alignment loss enforces consistency between the original and enhanced outputs using mean squared error (MSE):
\begin{equation}
    L_{total} = L_{Seg} + L_{Align}
\end{equation}

\section{Experiments}
\subsection{Implementation Details}

The proposed approach is implemented using PyTorch. The Adam optimizer is employed to update the network parameters. The input images are resized to 352 × 352 pixels using the bilinear interpolation. The initial learning rate is set to 8e-5 and a weight decay of 0.1 is used in network training. The batch size is set to 16 and the number of maximum epochs is set to 100. It takes 6 hours to about accomplish the training process on a GeForce GTX Titan X GPU.

\subsection{Datasets and Evaluation Metrics}
We evaluate our method on two benchmark datasets: MAS3K \cite{ECDNet}, which contains 3,103 images across seven superclasses, and RMAS \cite{masnet}, which consists of 3,014 images categorized into five superclasses.
Following \cite{ECDNet}, we adopt five widely used evaluation metrics for image segmentation, namely mean Intersection-over-Union ($mIoU$), Enhanced-alignment measure ($E_\phi$), Structure-measure ($S_\alpha$), weighted F-measure ($F_\beta^\omega$), and Mean Absolute Error (MAE).

\subsection{Comparison with State-of-the-Art Methods}
We compared our proposed method with 15 state-of-the-art models, including U2Net\cite{qin2020u2},
SINet\cite{fan2021concealed},
PFNet\cite{PFNet},
C2FNet\cite{C2FNet},
ECDNet\cite{ECDNet},
OCENet\cite{liu2022modelingOCENet},
ZoomNet\cite{pang2022zoom},
FDNet\cite{FDNet},
MASNet\cite{masnet},
H2Former\cite{he2023h2former},
MVGNet\cite{MVGNet},
CamoFocus\cite{khan2024camofocus},
SAM-Adapter\cite{chen2023sam},
SAM-DADF\cite{SAM-DADF},
MAS-SAM\cite{MAS-SAM}.

\subsubsection{Quantitative Comparison}
The performance of APGNet is compared with 15 state-of-the-art models using five evaluation metrics, as shown in Table \ref{tab1}. APGNet consistently achieves leading performance on both datasets across all five metrics, demonstrating  superiority in overall integrity, structural accuracy, and pixel-level precision. Specifically, on the MAS3K dataset, APGNet achieves a 3.37\% improvement in $mIoU$, a 3.13\% improvement in $S_\alpha$, a 4.31\% improvement in $F^\omega _\beta$, and a 2.48\% improvement in $E_\phi$ compared to MASNet. These results further validate the superior performance of APGNet in  statistical and analytical evaluations. Additionally, when compared to SAM-based fine-tuning methods, APGNet achieves strong performance on the RMAS dataset, particularly in the $S_\alpha$ and $F^\omega _\beta$ metrics, achieving 0.872 and 0.822, respectively. These results validate APGNet as an effective and innovative solution for marine animal segmentation in challenging underwater environments.

\begin{table}[h]
\caption{Five evaluation metrics are employed in this study, namely $mIoU\uparrow$, $S_\alpha\uparrow$,$F^\omega _\beta\uparrow$, $E_\phi\uparrow$, and $M\downarrow$. The symbols “↑” and “↓” indicate that larger and smaller values are better, respectively. The best results are highlighted in bold. }\label{tab1}
\renewcommand{\arraystretch}{1.1}
\centering
\scalebox{0.6}{
\begin{tabular}{c|c|ccccc|ccccc}
\hline
\multirow{2}{*}{Method} & \multirow{2}{*}{Year} & \multicolumn{5}{c|}{MAS3K}                             & \multicolumn{5}{c}{RMAS}              \\ \cline{3-12} 
                        &                       & $mIoU$  & $S_\alpha$     & $F^\omega _\beta$     & $E_\phi$     & \multicolumn{1}{c|}{$M$} & $mIoU$  & $S_\alpha$     & $F^\omega _\beta$     & $E_\phi$     & $M$     \\ \hline
U2Net                   & 2020                  & 0.654 & 0.812 & 0.711 & 0.851 & 0.047                  & 0.676 & 0.83  & 0.762 & 0.904 & 0.029 \\ \cline{1-12}
SINet                   & 2020                  & 0.658 & 0.820  & 0.725 & 0.884 & 0.039                  & 0.684 & 0.835 & 0.78  & 0.908 & 0.025 \\ \cline{1-12}
PFNet                   & 2021                  & 0.695 & 0.839 & 0.746 & 0.890  & 0.039                  & 0.694 & 0.843 & 0.771 & 0.922 & 0.026 \\ \cline{1-12}
C2FNet                  & 2021                  & 0.717 & 0.851 & 0.761 & 0.894 & 0.038                  & 0.721 & 0.858 & 0.788 & 0.923 & 0.026 \\ \cline{1-12}
ECDNet                  & 2021                  & 0.711 & 0.850  & 0.766 & 0.901 & 0.036                  & 0.664 & 0.823 & 0.689 & 0.854 & 0.036 \\ \cline{1-12}
OCENet                  & 2022                  & 0.667 & 0.824 & 0.703 & 0.868 & 0.052                  & 0.680  & 0.836 & 0.752 & 0.900   & 0.030  \\ \cline{1-12}
ZoomNet                 & 2022                  & 0.736 & 0.862 & 0.780  & 0.898 & 0.032                  & 0.728 & 0.855 & 0.795 & 0.915 & 0.022 \\ \cline{1-12}
FDNet                   & 2023                  & 0.726 & 0.852 & 0.781 & 0.902 & 0.033                  & 0.711 & 0.835 & 0.766 & 0.892 & 0.029 \\ \cline{1-12}
MASNet                  & 2023                  & 0.742 & 0.864 & 0.788 & 0.906 & 0.032                  & 0.731 & 0.862 & 0.801 & 0.92  & 0.024 \\ \cline{1-12}
H2Former                & 2023                  & 0.748 & 0.865 & 0.810  & 0.925 & 0.028                  & 0.717 & 0.844 & 0.799 & 0.931 & 0.023 \\ \cline{1-12}
MVGNet                  & 2024                  & 0.713 & 0.835 & 0.781 & 0.892 & 0.033                  & 0.698 & 0.835 & 0.762 & 0.899 & 0.026 \\ \cline{1-12}
CamoFocus               & 2024                  & 0.742 & 0.856 & 0.785 & 0.901 & 0.031                  & 0.713 & 0.845 & 0.772 & 0.909 & 0.025 \\ \cline{1-12}
SAM-Adapter             & 2023                  & 0.714 & 0.847 & 0.782 & 0.914 & 0.033                  & 0.656 & 0.816 & 0.752 & 0.927 & 0.027 \\ \cline{1-12}
SAM-DADF                & 2023                  & 0.742 & 0.866 & 0.806 & 0.925 & 0.028                  & 0.686 & 0.833 & 0.78  & 0.926 & 0.024 \\ \cline{1-12}
MAS-SAM                 & 2024                  & \textbf{0.788} & \textit{0.887} & \textbf{0.840}  & \textbf{0.938} & \textbf{0.025}                  & \textbf{0.742} & \textit{0.865} & \textit{0.819} & \textbf{0.948} & \textbf{0.021} \\ \cline{1-12}
Ours                    & -                     & \textit{0.767} & \textbf{0.891} & \textit{0.822} & \textit{0.929} & \textit{0.026}                  & \textbf{0.742} & \textbf{0.872} & \textbf{0.822} & \textit{0.936} & \textbf{0.021} \\ \cline{1-12}
\end{tabular}
}
\end{table}

\subsubsection{Visual Comparison}
\begin{figure}
  \centering
  \includegraphics[width=\linewidth]{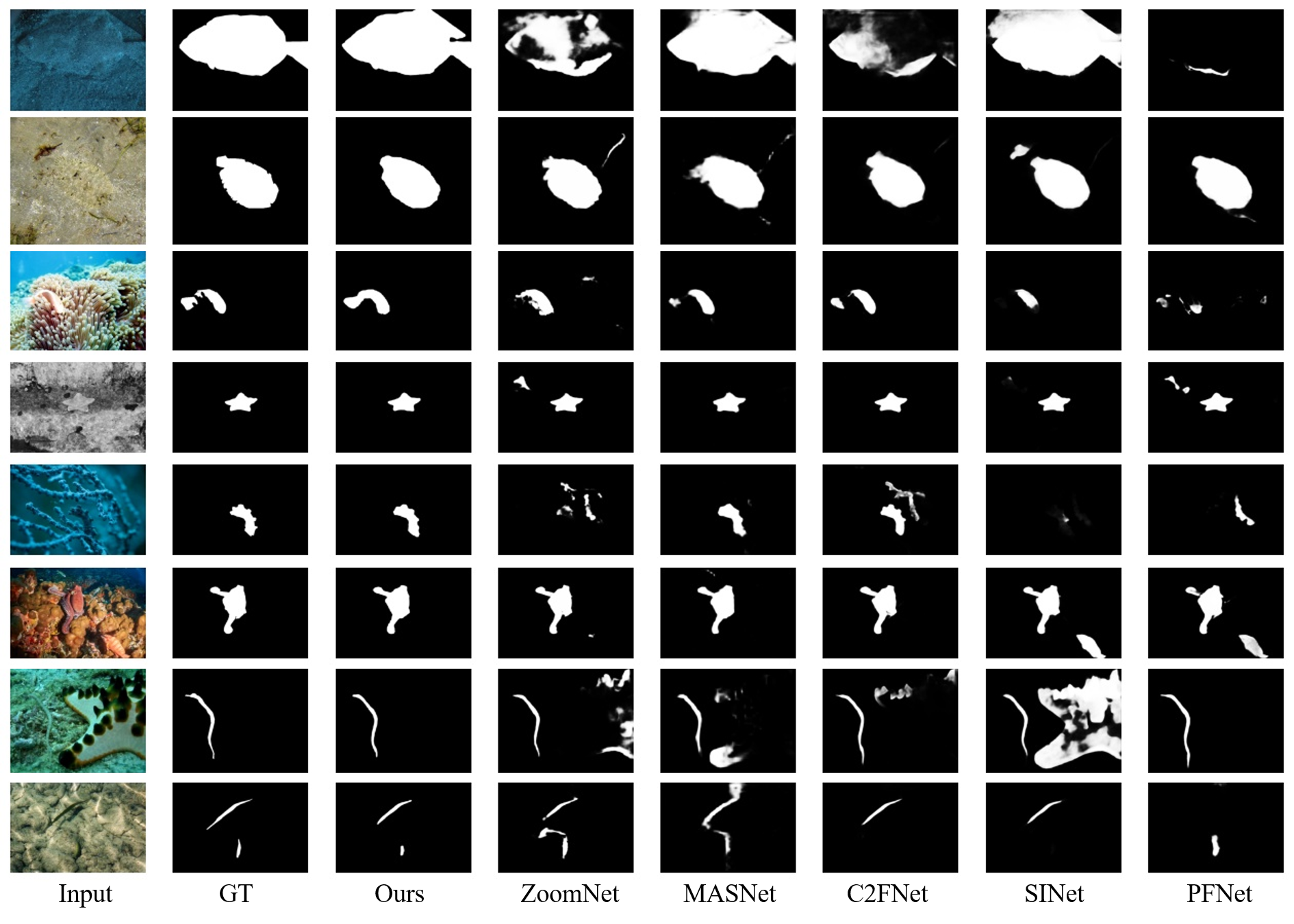}
  \caption{Visual comparison with other methods} \label{fig6}
\end{figure}

To intuitively demonstrate the superiority of the proposed method, Fig. \ref{fig6} presents a visual comparison between APGNet and other methods. As shown in the figure, APGNet more accurately identifies camouflaged objects in degraded underwater scenes, achieving precise localization (rows 1-2) and richer detail preservation (rows 3-6). Additionally, APGNet demonstrates good detection accuracy for elongated camouflaged objects commonly found in underwater scenes (rows 7-8). This improvement is attributed not only to the Siamese network and data augmentation strategy, which enhance the model’s ability to learn robust features, but also to the two-stage adaptive prior-guided strategy, which refines details and improves detection accuracy.

\subsection{Ablation Studies}
This section presents ablation experiments to evaluate the contributions of key components, including the Siamese network, data augmentation, the multi-scale progressive decoder (MPD), and the adaptive prior-guided module (APG). Five models are compared, including M1 (Baseline: PVTv2 with an ERF module), M2 (Baseline + MPD), M3 (Baseline + MPD + APG), M4 (M3 with data augmentation), and M5 (the complete model with Siamese integration). Table\ref{tab2} summarizes the quantitative results on two benchmark datasets, providing strong evidence of each component’s impact.
\begin{table}[h]
\caption{Quantitative evaluation for ablation studies on two datasets. The best results are highlighted in \textbf{bold}.}
\label{tab2}
\renewcommand{\arraystretch}{1.1}
\centering
\scalebox{0.65}{
\begin{tabular}{c|ccccc|ccccc}
\hline
\multirow{2}{*}{Method} & \multicolumn{5}{c|}{MAS3K} & \multicolumn{5}{c}{RMAS} \\ \cline{2-11}
                        & $mIoU$ & $S_\alpha$ & $F^\omega_\beta$ & $E_\phi$ & $M$ & $mIoU$ & $S_\alpha$ & $F^\omega_\beta$ & $E_\phi$ & $M$ \\ \hline
M1                      & 0.701  & 0.832      & 0.755            & 0.847    & 0.041 & 0.659  & 0.810      & 0.733            & 0.853    & 0.039 \\ \hline
M2                      & 0.714  & 0.849      & 0.773            & 0.861    & 0.037 & 0.686  & 0.831      & 0.749            & 0.872    & 0.035 \\ \hline
M3                      & 0.742  & 0.878      & 0.802            & 0.897    & 0.032 & 0.716  & 0.855      & 0.783            & 0.902    & 0.028 \\ \hline
M4                      & 0.712  & 0.851      & 0.764            & 0.851    & 0.038 & 0.662  & 0.835      & 0.742            & 0.868    & 0.035 \\ \hline
M5 (Ours)               & \textbf{0.767} & \textbf{0.891} & \textbf{0.822} & \textbf{0.929} & \textbf{0.026} & \textbf{0.742} & \textbf{0.872} & \textbf{0.822} & \textbf{0.936} & \textbf{0.021} \\ \hline
\end{tabular}
}
\end{table}

\textbf{Effectiveness of the Siamese Network Structure.} 
This section examines the impact of the Siamese network on model performance by removing it and applying only data augmentation. As shown in Table~\ref{tab2}, introducing the Siamese network, combined with alignment loss, encourages the model to learn shared semantic features by jointly supervising predictions from the original and augmented images. This significantly improves detection accuracy, outperforming the non-Siamese model across multiple metrics. Additionally, the Siamese network enhances feature alignment across different data distributions, increasing robustness in complex scenes.

\textbf{Effectiveness of the Data Augmentation Strategy.} 
The role of data augmentation is analyzed by comparing M3, M4, and M5 in Table~\ref{tab2}. Results show that applying data augmentation alone degrades performance due to distribution mismatch between training and testing data. Although augmentation enhances image quality, it alters the input distribution, leading to inconsistencies during inference. However, when integrated with the Siamese network, data augmentation mitigates this issue, leveraging the network’s structure to improve detection accuracy.

\textbf{Effectiveness of the Adaptive Prior-Guided Strategy.} 
This section evaluates the impact of the adaptive prior-guided strategy by comparing M2 and M3 in Table~\ref{tab2}. Results show improvements of 4.37\% in mIoU, 2.89\% in $S_\alpha$, 4.54\% in $F^\omega_\beta$, and 3.44\% in $E_\phi$, significantly enhancing camouflaged object detection. This demonstrates that the adaptive prior-guided strategy effectively leverages prior information, improving feature learning robustness and accuracy.

\section{Conclusion}
In this paper, we proposed APGNet, a novel architecture for underwater camouflaged object detection that effectively addresses the challenges of image degradation and natural camouflage in marine environments. By integrating a Siamese segmentation framework with Retinex-based data augmentation, a multi-scale progressive decoder, and an adaptive prior-guided module, APGNet leverages both position and boundary priors to refine segmentation progressively. Extensive experiments on benchmark datasets demonstrate that APGNet outperforms 15 state-of-the-art methods across multiple evaluation metrics, validating its superiority in overall integrity, structural accuracy, and pixel-level precision. 
In the future, we will investigate compressing our model into a lightweight model suitable for mobile devices and improving its efficiency in real-time applications.

\bibliographystyle{ACM-Reference-Format}
\bibliography{reference}
\end{document}